\documentclass[10pt,twocolumn,letterpaper]{article}

\usepackage{iccv}              
\usepackage{marvosym}

\definecolor{iccvblue}{rgb}{0.21,0.49,0.74}
\usepackage[pagebackref,breaklinks,colorlinks,allcolors=iccvblue]{hyperref}
\usepackage{tikz}

\def\paperID{*****} 
\def\confName{ICCV}
\def\confYear{2025}

\title{ViSE: A Systematic Approach to Vision-Only Street-View Extrapolation}

\author{
~~~~~~KaiyuanTan
~~~~~~Yingying Shen
~~~~~~Haiyang Sun$^\dagger$
~~~~~~Bing Wang \\
~~~~~~Guang Chen
~~~~~~Hangjun Ye\textsuperscript{\Letter}
\\
~ ~ Xiaomi EV
}
\begin{document}
\maketitle
\begin{abstract}
Realistic view extrapolation is critical for closed-loop simulation in autonomous driving, yet it remains a significant challenge for current Novel View Synthesis (NVS) methods, which often produce distorted and inconsistent images beyond the original trajectory. This report presents our winning solution which ctook first place in the RealADSim Workshop NVS track at ICCV 2025. To address the core challenges of street view extrapolation, we introduce a comprehensive four-stage pipeline. First, we employ a data-driven initialization strategy to generate a robust pseudo-LiDAR point cloud, avoiding local minima. Second, we inject strong geometric priors by modeling the road surface with a novel dimension-reduced SDF termed 2D-SDF. Third, we leverage a generative prior to create pseudo ground truth for extrapolated viewpoints, providing auxilary supervision. Finally, a data-driven adaptation network removes time-specific artifacts. On the RealADSim-NVS benchmark, our method achieves a final score of 0.441, ranking first among all participants.

\end{abstract} 
\let\thefootnote\relax
\footnotetext{
\small
\textsuperscript{$\dagger$} Project leader~~\textsuperscript{\Letter} Corresponding Author
}
\section{Introduction}
\label{sec:intro}
Validating autonomous driving algorithms in the real world faces high cost and significant safety risks, making high-fidelity simulation a critical enabling technology for advancing AD systems. Traditional game-engine-based simulators support closed-loop testing but suffer from a substantial domain gap and require labor-intensive, manual scene construction. Conversely, replay from real-world driving logs offers high visual fidelity but is limited to pre-recorded scenarios, unable to support interactive, closed-loop evaluation.

Novel View Synthesis (NVS) presents a compelling pathway to bridge this gap, aiming to construct interactive, 4D driving environments directly from real-world captures. However, a critical obstacle remains: the inherent sparsity of driving logs means that most NVS methods excel at view interpolation but struggle severely with view extrapolation. Under extrapolated viewpoints, current techniques based on volumetric primitives like NeRF\cite{nerf} or 3D Gaussian Splatting (3DGS)\cite{3dgs} often exhibit major geometric distortions and a collapse in texture realism.

The RealADSim challenge was established to directly address this critical gap. It provides a set of multi-traversal driving logs with a quantitative metric designed for evaluating extrapolation performance. 

In this report, we introduce a holistic pipeline that achieves robust and geometrically consistent street view extrapolation. Our solution systematically tackles the core challenges through four key contributions:

\begin{enumerate}
    \item A robust, LiDAR-free initialization strategy that generates a vision-based pseudo point cloud to provide a strong geometric prior and prevent overfitting.
    \item A dimension-reduced 2D Signed Distance Function (2D-SDF) that enforces a locally planar prior for road surfaces, improving geometric consistency under large viewpoint changes.
    \item An iterative pseudo-ground-truth framework that leverages a generative prior to provide supervision for unobserved, extrapolated viewpoints, effectively repairing artifacts.
    \item A data-driven time-invariance adaptation network that removes transient, time-specific features, ensuring robust performance across logs captured under different conditions.
\end{enumerate}
On the RealADSim benchmark, our team achieves a final score of 0.441, ranking 1st among all participants. 

\section{Preliminary}
\label{sec:method_preliminary}
\begin{figure*}
    \centering
    \includegraphics[width=1\linewidth]{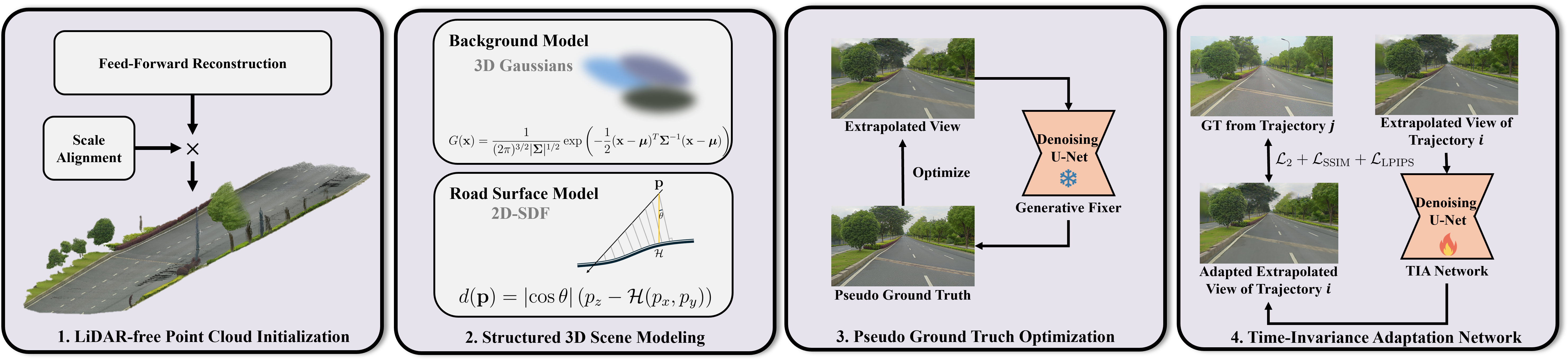}
    \caption{\textbf{Proposed Pipeline.} We introduce a four-stage framework for consistent novel-view synthesis. The process begins with (1) generating a robust, vision-based point cloud for initialization. This enables (2) detailed 3D reconstruction, where a 2D-SDF provides strong geometric priors for the road. To address remaining artifacts in unobserved areas, we employ (3) an iterative pseudo-GT strategy using a generative prior. Finally, (4) a time-invariance adaptation network ensures the output is temporally invariant across different driving logs.}
    \label{fig:pipeline}
\end{figure*}
\subsection{3D Gaussian Splatting}
\label{sec:3dgs_pre}

3D Gaussian Splatting (3DGS) \cite{3dgs} represents scenes using 3D Gaussian primitives. Each primitive is defined by a position $\mathbf{\mu} \in \mathbb{R}^{3}$, covariance matrix $\mathbf{\Sigma}\in \mathbb{R}^{3 \times 3}$, and opacity $\alpha\in\mathbb{R}^{+}$:
$
\mathcal{G}(\mathbf{x}) = \alpha \exp \left(-\frac{1}{2} (\mathbf{x}-\mathbf{\mu})^T \mathbf{\Sigma}^{-1} (\mathbf{x}-\mathbf{\mu}) \right)
$
The covariance matrix is factorized into scaling and rotation components: $\mathbf{\Sigma} = \mathbf{R}\mathbf{S}\mathbf{S}^\top\mathbf{R}^\top$.

For rendering, 3D Gaussians are transformed to camera coordinates via world-to-camera matrix $\mathbf{W}$ and projected to 2D using local affine approximation $\mathbf{J}$: $\mathbf{\Sigma}^{'} = \mathbf{J} \mathbf{W} \mathbf{\Sigma} \mathbf{W}^\top \mathbf{J}^\top$. By discarding the third dimension, we obtain 2D Gaussians $\mathcal{G}^{2D}$ with covariance $\mathbf{\Sigma}^{2D}$. The final color is computed via front-to-back alpha blending:
\begin{equation}
\mathbf{c}(\mathbf{x}) = \sum^K_{k=1} \mathbf{c}_k,\alpha_k,\mathcal{G}^{2D}k(\mathbf{x}) \prod{j=1}^{k-1} (1 - \alpha_j,\mathcal{G}^{2D}_j(\mathbf{x}))
\end{equation}
where $\mathbf{c}_k$ represents view-dependent color. This fully differentiable pipeline enables optimization through standard photometric losses.

\subsection{Rendering Signed Distance Functions}
\label{sec:sdf_pre}
 A common approach to representing a surface is through the zero-level set of a Signed Distance Function (SDF), where the function's value indicates the distance to the nearest surface, with the sign denoting whether the point lies inside or outside the surface. NeuS~\cite{neus} introduced a volume rendering technique for optimizing neural SDFs from posed images. 

Given a ray $\mathbf{r}$, with sampled points ${\mathbf{p}(t_i)}_{i=0}^N$, the opacity $\alpha_i$ at location $\mathbf{p}(t_i)$ is computed as:
\begin{equation}
\label{eq:neus}
    \alpha_i = \max \left (\frac{\Phi_s(d(\mathbf{p}(t_{i}))) - \Phi_s(d(\mathbf{p}(t_{i+1})))}{\Phi_s(d(\mathbf{p}(t_i)))}, 0 \right)
\end{equation}
Where $\Phi_s(x)=1/(1+\exp(-sx))$ is the Sigmoid-based cumulative distribution function, with $s$ controlling the sharpness of the transition and typically treated as a learnable parameter. Then, the color of each pixel is rendered using alpha-blending:
\begin{equation}
    \mathbf{C}(\mathbf{r}) = \sum_{i = 1}^{N} T_i\alpha_i\mathbf{c}_i, \ \ T_i=\prod_{j=1}^{i - 1} (1-\alpha_j)
\end{equation}

\section{Method}
\label{sec:method}

\paragraph{Problem Formulation}
Given a driving log comprising images ${I_i}_{i=1}^N$ and their associated ego-vehicle poses ${P_i}_{i=1}^N$, our objective is to learn a 3D scene representation that maps a 6-DoF pose to an image, $\mathcal{S}: SE(3)\rightarrow \mathbb{R}^{H\times W\times 3}$. This work focuses on the challenging task of synthesizing novel views from extrapolated viewpoints beyond the original trajectory.

We address four core challenges through a systematic, four-stage pipeline, illustrated in Figure~\ref{fig:pipeline}.

\subsection{3D Scene Initialization without LiDAR}
The absence of LiDAR data poses a significant challenge for initializing 3D Gaussians. Conventional Structure-from-Motion (SfM) \cite{sfm} points, often combined with random initialization, typically yield poor geometry due to the sparsity and limited viewing directions in driving logs. This frequently leads to convergence in a local minimum that fails under viewpoint extrapolation.

To overcome this, we generate a vision-based pseudo-LiDAR point cloud for robust initialization. We adopt VGGT \cite{vggt} for its favorable balance of quality and simplicity. As this method lacks absolute scale, we recover it by aligning the predicted camera poses with the ground-truth poses. The recovered scale is then multiplied to the predicted depth maps, which are subsequently unprojected and aggregated using the ground-truth poses to form a unified point cloud.

Although this pseudo-LiDAR point cloud can be noisy, it provides critical geometric initialization that enables faster 3DGS convergence, recovers finer details, and prevents implausible geometric distortions during extrapolation.

\subsection{Geometry-Aware 3D Scene Reconstruction}

\paragraph{2D-SDF for Road Surface Representation}
Accurately modeling the road surface is paramount for driving simulation, as it contains critical visual elements like road signs and lane markings. However, this is challenging due to weak textures and large viewpoint changes in extrapolated scenarios. Methods based on 3DGS \cite{3dgs} and NeRF \cite{nerf} often render severe distortions for the road surface under extrapolated trajectories.

While several methods~\cite{hugsim, autosplat, recondreamer++} explored constraining 3D Gaussians for a flat road surface, we consider an alternative representation. We introduce a novel 2D Signed Distance Function (2D-SDF) that incorporates the strong geometric prior that road surfaces are \textit{locally planar} and exhibit\textit{ globally smooth slope transitions}. This allows us to compress the representation from a general 3D SDF to two efficient 2D fields defined over the horizontal plane, modeling surface height and local slope, respectively. Under the local planar assumption, the signed distance from a 3D point $\mathbf{p}$ simplifies to:
\begin{equation}
d(\mathbf{p}) = |\cos\theta|\left(p_{z} - \mathcal{H}(p_{x},p_{y})\right),
\end{equation}
where $\mathcal{H}: \mathbb{R}^2 \rightarrow \mathbb{R}$ represents the surface height at horizontal coordinates $(p_x, p_y)$, and $\theta$ is the angle between the surface normal and the vertical axis. Both $\mathcal{H}$ and $|\cos\theta|$ depend solely on $(p_x, p_y)$ and are parameterized using neural networks. This leads to our 2D-SDF formulation\footnote{We use $\text{MLP}$ to denote networks for brevity; in practice, we use multi-level hash grids as in \cite{neuralangelo}.}:
\begin{equation}\label{eq:2dsdf}
\left\{
\begin{aligned}
d(\mathbf{p}) &= \text{MLP}_{\text{slope}}(\mathbf{p})\left[p_z -\text{MLP}_{\text{elevation}}(\mathbf{p})\right]\\
\mathbf{c}(\mathbf{p}, \mathbf{v}) &= \text{MLP}_{\text{color}}(\mathbf{p},\mathcal{F}(\mathbf{v}))
\end{aligned}
\right.
\end{equation}

Constrained to produce a smooth surface, the 2D-SDF ensures strong geometric consistency under extrapolation. Note that This dimension-reduced SDF also improves optimization efficiency over vanilla SDFs, converging in approximately 15 minutes per scene. Following previous SDF-based rendering methods (Section~\ref{sec:sdf_pre}), the 2D-SDF is optimized alongside the rest of the scene using reconstruction losses alone.

\paragraph{Sky Modeling and Composition}
We represent all above-ground objects using 3D Gaussians, following prior work \cite{pvg, hugs, omnire, streetgs}. The sky, being infinitely far away, is represented by an optimizable environment texture map, as in \cite{pvg, omnire}. To composite these representations, we assume a layering order where the road surface lies behind all above-ground objects, and the sky lies behind the road. The final image is rendered as:
\begin{align}
I = O_\text{gs} I_\text{gs} + (1-O_\text{gs})O_\text{road}I_\text{road} + (1-O_\text{gs})(1-O_\text{road})I_\text{sky},
\end{align}
where $I_\text{gs},I_\text{road}, I_\text{sky}$
  are the rendered colors from the 3D Gaussians, the road model, and the sky environment map, respectively, and 
$O_\text{gs}, O_\text{road}$ are their accumulated opacities.

\subsection{Iterative Pseudo Ground Truth Generation}
While our 2D-SDF effectively regularizes the road, non-road objects (e.g., vegetation, curbs, buildings) lack a universal geometric prior. These objects often exhibit severe distortions or floating artifacts under extrapolation due to insufficient observations.

To provide explicit supervision for these unobserved regions, we employ an iterative pseudo ground truth (GT) generation strategy. Following recent work \cite{drivedreamer4d, streetcrafter, recondreamer, freevs, freesim, recondreamer++, drivex}, we leverage a pre-trained diffusion model as a generative fixer. We adopt the off-the-shelf Difix3D+ \cite{difix3d+} model for this purpose. At scheduled intervals during training, we synthesize pseudo-GT images for progressively extrapolated camera poses. These poses are generated by interpolating between the nearest training views and the target extrapolated test views. We first render the current 3D scene at these target poses and then use the generative fixer to refine the noisy renderings into pseudo-GT images, which are added back to the training set (see Figure~\ref{fig:pipeline}, Part 3).

The pseudo-GT supervision loss is defined as:
\begin{equation}
\mathcal{L}_{\text{pseudo}} = \lambda_{\text{LPIPS}} \cdot \mathcal{L}_{\text{LPIPS}} + \lambda_{\text{L1}} \cdot \mathcal{L}_{\text{L1}}.
\end{equation}

A naïve application of this supervision introduces a trade-off: while it effectively reduces floating artifacts (improving LPIPS), it can also hallucinate unrealistic details such as over-saturated textures or spurious objects, which harm pixel-level metrics (PSNR/SSIM). To balance this, we significantly down-weight $\lambda_{\text{L1}}$ relative to $\lambda_{\text{LPIPS}}$. This prioritizes perceptual quality while mitigating the impact of pixel-level inaccuracies in the pseudo-GT.

\subsection{Time-Invariance Adaptation Network}
A unique challenge in this competition is the need for extrapolation not only in space but also in time. Different lanes were recorded at different times, leading to variations in illumination, weather, and transient objects (e.g., puddles, shadows). A model that naïvely memorizes these time-specific details produces inconsistent results when rendering a scene in the style of a different log. We found that factoring out these time-varying artifacts is critical for achieving realistic and consistent view synthesis. For instance, time-specific elements like reflections on asphalt, puddles, and cloud patterns should be suppressed, while dynamic textures like foliage should be smoothed to avoid hallucinated details.

To achieve temporal invariance, we introduce a lightweight Time-Invariance Adaptation Network (TIA-Net). This network acts as a post-processing module that learns to remove time-specific artifacts from a rendered image:
\begin{align}
I' = \text{TIA}_\theta(I),
\end{align}
where $\theta$ represents the learnable parameters.

We train TIA-Net in a data-driven manner using multi-traversal data from the public portion of the Para-Lane~\cite{paralane} dataset. Specifically, we reconstruct a 3D scene from one trajectory (e.g. $\text{LOG}_i$) using our full pipeline. We then render this scene using the camera poses from a different trajectory, $\text{LOG}_j$, which was captured at a different time. The rendered images are passed through TIA-Net, and the output is compared to the real images from $\text{LOG}_j$.using photometric losses.

This objective forces TIA-Net to learn a mapping that effectively removes time-dependent features (e.g., specific lighting, reflections) while preserving the underlying scene structure, resulting in a consistent and generalizable appearance.

We train TIA-Net by fine-tuning the Difix3D+~\cite{difix3d+} model, as it provides a fast, one-step image-to-image refinement framework that is well-suited for this adaptation task.
\section{Experiments}

\begin{table}[htb]
\centering
\small
\begin{tabular}{llccc} 
\toprule
  Rank&Team& PSNR$\uparrow$&SSIM$\uparrow$& LPIPS$\downarrow$\\
  \cmidrule{1-5}
 1&XiaomiEV Team& \textbf{18.228}&\textbf{0.514}& \textbf{0.288}\\
 2&Qualcomm AI Research& 17.887&0.492&0.289\\
 3&R2D2&18.009&0.496&0.361\\
 4&MeowAndDoggy& 17.857&0.490&0.371\\
 5&aowei& 16.72&0.484& 0.401\\
\bottomrule
\end{tabular}
\caption{\textbf{Results on the RealADSim-NVS Benchmark.} We only copy results for the top-5 teams. The full results are available at https://huggingface.co/spaces/XDimLab/ICCV2025-RealADSim-NVS. }
\label{tab:main_results}
\end{table}
\subsection{Main Results}
We present results on the RealADSim-NVS benchmark, which is specifically designed to assess the challenging task of novel view extrapolation in autonomous driving scenarios. As shown in Table~\ref{tab:main_results}, our full pipeline, designated as \textbf{XiaomiEV Team}, achieves state-of-the-art performance, ranking first among all participants. This demonstrates the effectiveness of our structured vision-based pipeline in generating realistic and geometrically consistent images at extrapolated views.
\subsection{Ablation Study}
To validate the contribution of each component in our pipeline, we conduct a comprehensive ablation study. The results are summarized in Table~\ref{tab:ablation}
\begin{itemize}
\item[\textcircled{\raisebox{-0.9pt}{2}}] \textbf{2D-SDF Road Prior:} Introducing a strong geometric prior for the road surface via our 2D-SDF model provides a crucial foundation. It preserves PSNR/SSIM while significantly improving LPIPS (0.500 vs. 0.513). This indicates that the model successfully mitigates geometric distortions under extrapolation, leading to more structurally coherent and realistic outputs.
\item[\textcircled{\raisebox{-0.9pt}{3}}] \textbf{Pseudo-LiDAR Initialization:} Initializing 3DGS with a vision-based pseudo point cloud yields a further boost across all metrics. This step alleviates overfitting to the training viewpoints and helps the model escape poor local minima, resulting in more accurate geometry for both distant scenes and near-field objects.

\item[\textcircled{\raisebox{-0.9pt}{4}}] \textbf{Pseudo GT with Generative Prior:} This stage delivers a substantial leap in performance, most notably a ~20\% relative improvement in LPIPS (0.396 vs. 0.47). By providing explicit supervision for unobserved regions, the generative prior effectively ``inpaints'' plausible scene content and eliminates floating artifacts, dramatically enhancing the visual plausibility of extrapolated views.

\item[\textcircled{\raisebox{-0.9pt}{5}}] \textbf{TIA Network:} The final addition of our Time-Invariance Adaptation Network achieves highest improvement. By learning to remove transient, time-specific elements (e.g., illumination changes, shadows, and moving clouds), the TIA network produces a stable, time-agnostic scene representation. This is critical for robust performance under the spatio-temporal extrapolation required by the benchmark.
\end{itemize}

\begin{table}[htb]
\centering
\small
\begin{tabular}{llccc} 
\toprule
Exp. ID & Method & PSNR$\uparrow$&SSIM$\uparrow$& LPIPS$\downarrow$\\
\cmidrule{1-5}
\textcircled{\raisebox{-0.9pt}{1}}&baseline& 16.154&0.480& 0.513\\
\textcircled{\raisebox{-0.9pt}{2}}&\textcircled{\raisebox{-0.9pt}{1}} + 2D-SDF& 16.170&0.484&0.500\\
\textcircled{\raisebox{-0.9pt}{3}}&\textcircled{\raisebox{-0.9pt}{2}} + Pseudo LiDAR&16.369&0.493&0.47\\
\textcircled{\raisebox{-0.9pt}{4}}&\textcircled{\raisebox{-0.9pt}{3}} + Pseudo GT& 17.286&0.507&0.396\\
\textcircled{\raisebox{-0.9pt}{5}}&\textcircled{\raisebox{-0.9pt}{4}} + TIA Network& \textbf{18.228}&\textbf{0.514}& \textbf{0.288}\\
\bottomrule
\end{tabular}
\caption{\textbf{Ablation of Key Steps}}
\label{tab:ablation}
\end{table}
\section{Conclusion}
This report presents our solution for robust street view extrapolation that earned 1st place in the RealADSim-NVS challenge. Our method systematically addresses key challenges through LiDAR-free initialization, a novel 2D-SDF road prior, generative pseudo-GT supervision, and temporal adaptation. Extensive experiments validate each component's contribution to our state-of-the-art performance. 

{
    \small
    \bibliographystyle{ieeenat_fullname}
    \bibliography{main}
}


\end{document}